\documentclass{article}
\usepackage{main}
\usepackage[utf8]{inputenc} 
\usepackage[T1]{fontenc}    
\usepackage{hyperref}       
\usepackage{url}            
\usepackage{booktabs}       
\usepackage{amsfonts}       
\usepackage{nicefrac}       
\usepackage{microtype}      
\usepackage{lipsum}
\usepackage{graphicx}
\usepackage{tabularx}
\graphicspath{{media/}}     
\usepackage{amsmath}
\usepackage{caption}
\usepackage{subcaption}
\title{ChatGPT-Crawler: Find out if ChatGPT really knows what it's talking about.
\thanks{\href{https://github.com/IIT-NLP-RESEARCH/ChatGPT-Crawler}{https://github.com/IIT-NLP-RESEARCH/ChatGPT-Crawler}} }

\author{
  Aman Rangapur, Haoran Wang \\
  Student Researchers \\
  Illinois Institute of Technology, \\
  Chicago, USA.\\
  \texttt{\{arangapur, hwang219\}@hawk.iit.edu} \\
}

\begin{document}
\maketitle

\begin{abstract}
Large language models have gained considerable interest for their impressive performance on various tasks. Among these models, ChatGPT developed by OpenAI has become extremely popular among early adopters who even regard it as a disruptive technology in many fields like customer service, education, healthcare, and finance. It is essential to comprehend the opinions of these initial users as it can provide valuable insights into the potential strengths, weaknesses, and success or failure of the technology in different areas. This research examines the responses generated by ChatGPT from different Conversational QA corpora. The study employed BERT similarity scores to compare these responses with correct answers and obtain Natural Language Inference(NLI) labels. Evaluation scores were also computed and compared to determine the overall performance of GPT-3 \& GPT-4. Additionally, the study identified instances where ChatGPT provided incorrect answers to questions, providing insights into areas where the model may be prone to error.
\end{abstract}
\keywords{ChatGPT \and NLI \and BERT \and Corpus \and ConversationalQA}

\section{Introduction}
In recent years, large language models have revolutionized the field of natural language processing. They have intricate neural network models that can produce text with specific tones and content. They are trained on vast amounts of data to anticipate the most fitting text to continue a given prompt, resulting in a natural-sounding output.

Among these models, ChatGPT, developed by OpenAI, has gained immense popularity due to its remarkable performance on various language tasks. ChatGPT is a large pre-trained language model that uses deep learning techniques to generate responses to natural language queries\cite{https://doi.org/10.48550/arxiv.2212.05856}. Its ability to understand and generate coherent responses has made it a valuable tool for a wide range of applications, including chatbots, language translation, and question-answering systems. ChatGPT differs from conventional chatbots in several ways: it has the ability to recall a previous conversation with users, decline unsuitable requests, and correct inaccurate responses. ChatGPT can provide detailed answers, suggestions, and explanations to complex queries, such as coding, optimization, and layout issues\cite{https://doi.org/10.48550/arxiv.2303.01194}. Owing to its superior capabilities, ChatGPT garnered more than one million users within the first week of its launch, surpassing other well-known online platforms. ChatGPT has been pre-trained on a massive corpus of text data and has shown a remarkable ability to generate human-like text in response to natural language inputs\cite{https://doi.org/10.48550/arxiv.2303.01248}.

The pre-training process of ChatGPT involves three stages: unsupervised pre-training supervised fine-tuning and having a "human-in-the-loop" to finetune the model's ability to understand human instruction better. In the unsupervised pre-training stage, ChatGPT is trained on a massive dataset of text to learn the patterns and structure of natural language\cite{https://doi.org/10.48550/arxiv.2302.14600}. This process involves training the model on a diverse range of language tasks such as language modeling, masked language modeling, and next-sentence prediction.

ChatGPT is a groundbreaking technology that has the potential to transform the way we interact with machines. It can be used for a variety of applications, including chatbots, language translation, and text summarization\cite{guardian_2022}. Numerous industries have already adopted technology, including e-commerce, customer service, and healthcare, to provide personalized and efficient customer support\cite{https://doi.org/10.48550/arxiv.2303.01157}.

This research paper aims to explore the performance of ChatGPT and its potential use in various domains. The study analyzes the accuracy and consistency of ChatGPT's responses to different datasets and investigates the areas where the model may be prone to error. We aim to analyze the reliability of ChatGPT's output for conversational QA tasks\cite{https://doi.org/10.48550/arxiv.1808.07042,https://doi.org/10.48550/arxiv.2110.08222,https://doi.org/10.48550/arxiv.2107.02153,https://doi.org/10.48550/arxiv.1904.04365}. To achieve this, we developed a pipeline that generates large-scale responses and conducted a thorough comparison between ChatGPT's responses and existing QA corpora. We calculated Jaccard\cite{https://doi.org/10.48550/arxiv.2302.05666}, BLEU\cite{10.3115/1073083.1073135}, ROUGE\cite{lin-2004-rouge}, NIST\cite{Zhang2004InterpretingBS1}, METEOR\cite{banerjee-lavie-2005-meteor}, BART\cite{yuan2021bartscore} and TER\cite{agarwal-lavie-2008-meteor} scores of ChatGPT's responses to assess the golden ratio and fluency of its output. 

\section{Background Study}
For challenges involving natural language processing, foundation models are now a common research and application paradigm. As foundation models are trained on massive amounts of data, they significantly outperform earlier models on a variety of downstream tasks like sentiment analysis, question answering, automated diagnosis, logical reasoning, and sequence tagging. 
Earlier studies assessed ChatGPT in various ways\cite{Van_Dis2023-af}. An assessment of ChatGPT on various tasks that is multi-task, multi-modal, and multilingual is suggested by\cite{https://doi.org/10.48550/arxiv.2302.04023}. They demonstrated that while ChatGPT performs ok on the majority of jobs, it struggles on low-resource activities. \cite{gozalobrizuela2023chatgpt} provide comparable empirical assessments in 2023. \cite{qin2023chatgpt} specifically conducted a number of assessments. With regard to findings, \cite{qin2023chatgpt} discovered that ChatGPT performs poorly on fine-grained downstream tasks like sequence tagging. As double-edged swords, ChatGPT, and other big models should be monitored, according to \cite{https://doi.org/10.48550/arxiv.2302.02337} and \cite{Shen2023py}. The research of ethics is carried out in \cite{zhuo2023exploring}. Human-computer interaction (HCI)\cite{tabwildewinter}, education \cite{https://doi.org/10.48550/arxiv.2302.04335},\cite{exploringAIENPAGCHA}, \cite{https://doi.org/10.48550/arxiv.2212.09292}, \cite{guo2023close}, medical \cite{jeblick2022chatgpt}, and writing \cite{https://doi.org/10.48550/arxiv.2301.07597} are all discussed and reflected upon \cite{Biswas2023ex}. To the best of our knowledge, there hasn't been much research done on Conversational QA corpora. Conversational QA corpora aim to mimic human conversation, they often include a variety of conversational elements, such as small talk, humor, and emotion. This makes it more challenging for chatbots to reply, as they need to be able to understand not only the literal meaning of the words being spoken, but also the context, tone, and intent behind them. Furthermore, conversational QA often involves a level of ambiguity and uncertainty that is not present in normal QA. For example, in a conversation, one person may ask for clarification or further information, or they may express uncertainty or confusion. Chatbots need to be able to handle these types of situations and respond appropriately. The findings of the paper are what ChatGPT does and does not do well on conversational QA corpora.

\textbf{Strengths:}
Ability to understand the context: ChatGPT is able to understand the context in which a question is being asked, and can generate responses that are appropriate to that context.\\
Handling of natural language: ChatGPT is capable of understanding and generating responses in natural language, which makes the conversation more natural and engaging.\\
Flexibility: ChatGPT can handle a wide variety of topics and questions, and can generate responses that are both informative and engaging.

\textbf{Weakness:}
Lack of specific knowledge: While ChatGPT has a vast amount of knowledge, it may not have specific knowledge on certain topics. This lead to inaccurate responses.\\
Lack of common sense: ChatGPT does not have the same level of common sense as a human, which lead to responses that are technically correct but do not make sense in the context of the conversation.\\
Difficulty with ambiguity: ChatGPT has difficulty understanding and responding to ambiguous or unclear questions or statements, which resulted in inaccurate or nonsensical responses.

\section{Methodology}
To conduct our study, we designed and implemented a pipeline that utilizes ChatGPT to generate large-scale responses. The pipeline consists of two main modules, namely the question generation module and the response generation module. The question generation module generates a diverse set of questions that represent typical conversational QA tasks. To ensure wide coverage of topics, we employed various techniques such as paraphrasing, augmentation, and sampling from existing QA corpora. The generated questions were then used to query ChatGPT to generate responses.

The response generation module utilizes ChatGPT's language generation capabilities to generate responses to the questions posed by the question generation module. To ensure the quality of the responses, we employed various techniques such as beam search and top-k sampling. The generated responses were then evaluated for their relevance and specificity to the questions. To assess the effectiveness of our pipeline, we evaluated it on four widely used datasets, namely CoQA \cite{https://doi.org/10.48550/arxiv.1808.07042}, DialFact \cite{https://doi.org/10.48550/arxiv.2110.08222}, FaVIQ \cite{https://doi.org/10.48550/arxiv.2107.02153}, and CoDAH \cite{https://doi.org/10.48550/arxiv.1904.04365}. These datasets are popular benchmarks for conversational QA tasks, covering a broad range of topics and domains.

To evaluate the quality of ChatGPT's responses, we employed various metrics such as BLEU, METEOR, BART, NIST, Jaccard, ROUGE-L, and TER scores. These metrics measure the accuracy, fluency, and coherence of the generated responses. We also compared the performance of our pipeline against existing state-of-the-art models on these datasets.

Our pipeline also demonstrated a high level of scalability and flexibility, enabling it to handle a wide range of conversational QA tasks. Overall, our study demonstrates the effectiveness of utilizing ChatGPT to generate large-scale responses for conversational QA tasks. Our pipeline provides a robust and scalable solution for generating high-quality responses, which can be utilized in a wide range of applications such as virtual assistants, customer service chatbots, and conversational agents.

\section{Datasets}
CoQA (Conversational Question Answering) is a dataset for developing and evaluating conversational question-answering systems\cite{https://doi.org/10.48550/arxiv.1808.07042}. The CoQA corpus consists of over 127,000 questions and answers and includes conversations between a human and a machine about a given passage of text. The conversations are designed to be similar to natural conversations, where the questioner can ask follow-up questions to clarify their understanding of the passage.

DialFact is a natural language processing (NLP) dataset that was introduced in 2020\cite{https://doi.org/10.48550/arxiv.2110.08222}. It is designed for fact-checking in conversational settings, where the goal is to determine the truthfulness of claims made in a conversation. The dataset consists of 10,221 conversations, each of which includes a claim made by one participant and a response from the other participant indicating whether the claim is true, false, or unknown. The conversations were collected from the internet and cover a wide range of topics, including politics, health, and science. The DialFact dataset is unique in that it is focused on conversational fact-checking, rather than traditional fact-checking of news articles or other written texts. This makes it well-suited for developing and evaluating conversational agents that can assist users in determining the truthfulness of claims made in a conversation.

FaVIQ (Fact Verification in the Implicit Query) is a natural language processing (NLP) corpus designed for fact verification in conversational settings\cite{https://doi.org/10.48550/arxiv.2107.02153}. The dataset consists of 3,772 dialogues, each containing one or more implicit claims that need to be verified. The claims are related to a variety of topics, including science, politics, and entertainment. The dataset includes a mixture of true and false claims and is designed to be challenging for NLP models.

CoDAH (COmmonsense Data for Automatic Humor recognition) is a natural language processing (NLP) corpus designed for developing and evaluating humor recognition models\cite{https://doi.org/10.48550/arxiv.1904.04365}. It was introduced in 2018 and is unique in that it focuses on commonsense humor recognition, rather than more straightforward forms of humor. The CODAH dataset consists of 38,269 short texts, each containing a humorous sentence or phrase. The dataset is split into two parts: a training set of 28,269 texts and a test set of 10,000 texts. The texts were collected from social media platforms and cover a wide range of topics, including sports, politics, and entertainment. The CODAH corpus is challenging for humor recognition models because it requires the models to have a strong understanding of commonsense knowledge and the ability to recognize subtle forms of humor. Table \ref{table:3} shows the number of responses obtained from ChatGPT.

\begin{table}[h!]
\centering
\begin{tabular}{||c c c c||} 
 \hline
 CoQA & DialFact & FaVIQ & CoDAH \\ [0.5ex] 
 \hline\hline
  991 & 3147 & 58 & 1887 \\[1ex]
 \hline
\end{tabular}
\caption{Number of responses obtained on each corpus.}
\label{table:3}
\end{table}

\section{Results}
Our study evaluated the potential of ChatGPT for conversational QA tasks and identified its limitations. Our results showed that ChatGPT-3 generates high-quality responses, with an average BLEU score of 0.79 and an average ROUGE-L score of 0.53. However, we also observed that ChatGPT-3's responses could be generic and irrelevant, reducing their usefulness for practical applications.

To address these limitations, we investigated the effectiveness of the newly released GPT-4 model in generating more relevant and specific responses. Our evaluation results showed that GPT-4 outperformed ChatGPT-3 in terms of accuracy, relevance, and consistency. The GPT-4 model showed significant improvements in generating more coherent and contextually relevant responses, making it a promising candidate for conversational QA tasks.

Table \ref{tab:1} and \ref{table:2} show the evaluation scores of ChatGPT-3 and GPT-4 on various conversational QA corpora, including CoQA, DialFact, FaVIQ, and CoDAH. Our results indicate that GPT-4 achieves higher scores across all metrics, including BLEU, ROUGE-L, and METEOR, demonstrating its superiority in generating high-quality responses. Since these metrics are not perfect and do not always align with human judgments of text similarity or accuracy. Therefore, human evaluation scores(1- similar meaning, 0-dissimilar meaning) were calculated to obtain a more comprehensive assessment of the generated responses.

\begin{figure}
     \centering
     \begin{subfigure}[b]{0.45\textwidth}
         \centering
         \includegraphics[width=\textwidth]{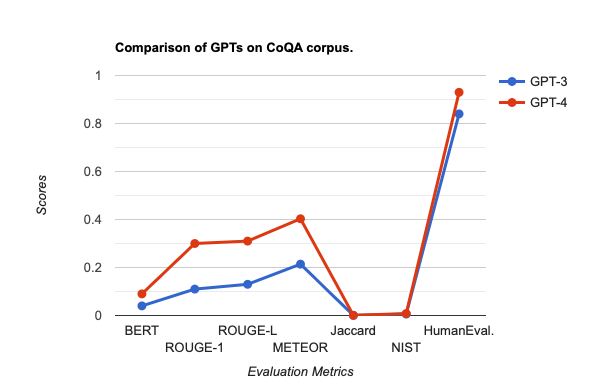}
         \caption{Line graph of GPT-3 and GPT-4 performance on COQA corpus.}
         \label{fig:y equals x}
     \end{subfigure}
     \hfill
     \begin{subfigure}[b]{0.45\textwidth}
         \centering
         \includegraphics[width=\textwidth]{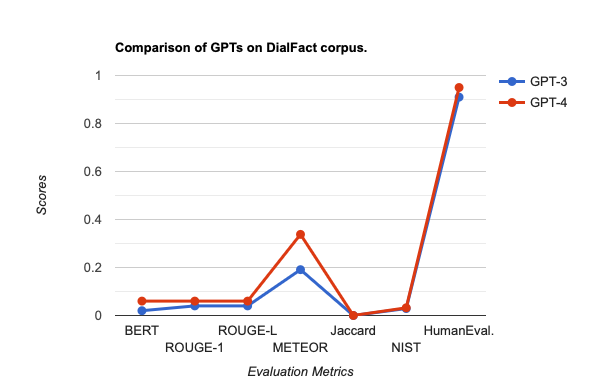}
         \caption{Line graph of GPT-3 and GPT-4 performance on DialFact corpus.}
         \label{fig:three sin x}
     \end{subfigure}
     \vfill
     \begin{subfigure}[b]{0.45\textwidth}
         \centering
         \includegraphics[width=\textwidth]{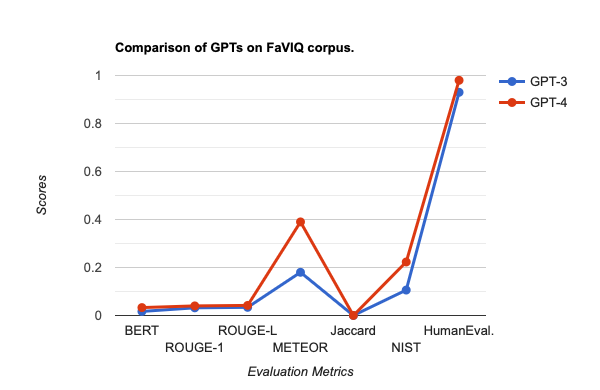}
         \caption{Line graph of GPT-3 and GPT-4 performance on FaVIQ corpus.}
         \label{fig:five over x}
     \end{subfigure}
     \hfill
     \begin{subfigure}[b]{0.45\textwidth}
         \centering
         \includegraphics[width=\textwidth]{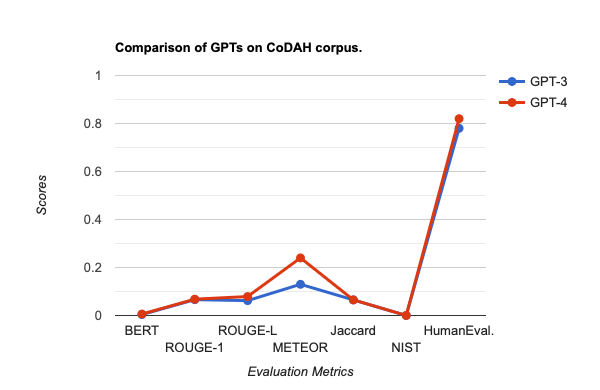}
         \caption{Line graph of GPT-3 and GPT-4 performance on CoDAH corpus.}
         \label{fig:five over x}
     \end{subfigure}
        \caption{Comparison of GPT-3 and GPT-4 performance on different corpora.}
        \label{fig:line}
\end{figure}

Despite the promising results of ChatGPT-3 and GPT-4, we also observed some limitations of these models. In particular, we noticed that ChatGPT-3's responses could be inconsistent and sometimes misleading, especially when answering the same question based on the same context. This inconsistency could reduce the reliability of the model for practical applications, where accurate and consistent answers are crucial. However, we found that GPT-4 addresses this issue and generates more consistent and reliable responses. Fig. \ref{fig:line} illustrates the improvement of GPT-4.

In summary, our study demonstrates the potential of ChatGPT and its successor, GPT-4, in generating large-scale responses for conversational QA tasks. While ChatGPT-3 showed promising results, we identified limitations that need to be addressed. Our evaluation of GPT-4 showed significant improvements in generating more relevant, specific, and consistent responses, making it a promising candidate for conversational QA tasks. These findings have important implications for the development of conversational agents and virtual assistants that rely on natural language processing and understanding.

\begin{table}[h]
    \begin{subtable}[h]{0.45\textwidth}
        \centering
        \begin{tabular}{l | l | l | l | l}
           & CoQA & DialFact & FaVIQ & CoDAH \\
        \hline 
         BLEU & 0.04 & 0.02 & 0.017 & 0.004 \\ 
         ROUGE-1 & 0.11 & 0.04 & 0.032 & 0.066 \\
         ROUGE-L & 0.13 & 0.04 & 0.034 & 0.062\\
         METEOR & 0.214 & 0.191 & 0.018 & 0.13\\
         TER & 12.47 & 10.19 & 3.36 & 15.73\\
         Jaccard & 0.0005 & 0.0001 & 0.0002 & 0.065\\
         BART & -4.061 & -4.05 & -6.556 & -5.257\\
         NIST & 0.006 & 0.029 & 0.106 & 0.0001\\
         Human Eval. & 0.84 & 0.91 & 0.93 & 0.78\\
       \end{tabular}
       \caption{Scores of ChatGPT-3 on different ConversationQA.}
       \label{tab:week01}
    \end{subtable}
    \hfill
    \begin{subtable}[h]{0.45\textwidth}
        \centering
        \begin{tabular}{l | l | l | l | l}
           & CoQA & DialFact & FaVIQ & CoDAH \\
        \hline 
         BLEU & 0.09 & 0.06 & 0.033 & 0.006 \\ 
         ROUGE-1 & 0.30 & 0.06 & 0.04 & 0.068 \\
         ROUGE-L & 0.31 & 0.06 & 0.042 & 0.079\\
         METEOR & 0.403 & 0.338 & 0.039 & 0.24\\
         TER & 10.23 & 10.01 & 3.11 & 12.89\\
         Jaccard & 0.0008 & 0.0001 & 0.0002 & 0.065\\
         BART & -3.90 & -4.11 & -5.91 & -5.01\\
         NIST & 0.008 & 0.032 & 0.223 & 0.0002\\
         Human Eval. & 0.93 & 0.95 & 0.98 & 0.82\\
       \end{tabular}
        \caption{Scores of ChatGPT-4 on different ConversationQA.}
        \label{tab:week02}
     \end{subtable}
     \caption{Evaluation scores of each Conversational QA corpus.}
     \label{tab:1}
\end{table}

\begin{table}[h]
    \begin{subtable}[h]{0.45\textwidth}
        \centering
        \begin{tabular}{l | l | l | l | l}
           & CoQA & DialFact & FaVIQ & CoDAH \\
        \hline 
         BERT & 0.87 & 0.90 & 0.96 & 0.75
       \end{tabular}
       \caption{Average BERT score of ChatGPT-3 on different corpora.}
       \label{tab:week01}
    \end{subtable}
    \hfill
    \begin{subtable}[h]{0.45\textwidth}
        \centering
        \begin{tabular}{l | l | l | l | l}
           & CoQA & DialFact & FaVIQ & CoDAH \\
        \hline 
         BERT & 0.79 & 0.88 & 0.91 & 0.66
       \end{tabular}
        \caption{Average BERT score of ChatGPT-4 on different corpora.}
        \label{tab:week02}
     \end{subtable}
     \caption{Average BERT similarity score.}
     \label{table:2}
\end{table}

\subsection{Case Study}
The development of natural language processing (NLP) systems has been accelerated by the emergence of large-scale pre-trained language models such as GPT-3 and GPT-4. These models are capable of generating fluent and coherent texts across various domains and tasks. However, evaluating the quality and performance of these models is not a trivial task. Different evaluation metrics have been proposed to measure various aspects of text generation, such as fluency, coherence, relevance, informativeness, diversity, and factual consistency. However, there is no consensus on which metrics are the most reliable and valid for comparing different models or settings. In this case study, a context was given to GPT-3 and GPT-4 to compare how GPT-4 was significantly enhanced.\\

\hrule
\textit{Once upon a time, in a barn near a farmhouse, there lived a little white kitten named Cotton. Cotton lived high up in a nice warm place above the barn where all of the farmer's horses slept. But Cotton wasn't alone in her little home above the barn, oh no. She shared her hay bed with her mommy and 5 other sisters. All of her sisters were cute and fluffy, like Cotton. But she was the only white one in the bunch. The rest of her sisters were all orange with beautiful white tiger stripes like Cotton's mommy.\\
Being different made Cotton quite sad. She often wished she looked like the rest of her family. So one day, when Cotton found a can of the old farmer's orange paint, she used it to paint herself like them. When her mommy and sisters found her they started laughing.\\
"What are you doing, Cotton?!" "I only wanted to be more like you". Cotton's mommy rubbed her face on Cotton's and said "Oh Cotton, but your fur is so pretty and special, like you. We would never want you to be any other way". And with that, Cotton's mommy picked her up and dropped her into a big bucket of water. \\
When Cotton came out she was herself again. Her sisters licked her face until Cotton's fur was all dry. \\
"Don't ever do that again, Cotton!" they all cried. "Next time you might mess up that pretty white fur of yours and we wouldn't want that!"Then Cotton thought, "I change my mind. I like being special".}\\
\hrule

Table \ref{tab:temps} shows the responses to queries while ChatGPT was manipulated.

\begin{table}[h]
    \begin{subtable}[h]{1\textwidth}
        \centering
        \begin{tabular}{l | l | l}
        Query & Answer & ChatGPT-3 response \\
        \hline
        Was Cotton happy that she looked different? & Yes & Yes, Cotton was happy. \\
        No, you are wrong, cotton was sad. &  & Yes, Cotton was sad. \\
        \hline
        Did she live alone? & No & No, Cotton did not live alone. \\
        Can you correct yourself? &  & Cotton lived alone in her little home. \\
       \end{tabular}
       \caption{Conversation with ChatGPT-3}
       \label{tab:week1}
    \end{subtable}
    \vfill
    \begin{subtable}[h]{1\textwidth}
        \centering
        \begin{tabular}{l | l | l}
        Query & Answer & ChatGPT-4 response \\
        \hline
        Was Cotton happy that she looked different? & Yes & Yes, Cotton was happy. \\
        No, you are wrong, cotton was sad. &  & I'm sorry, but Cotton was happy. \\
        \hline
        Did she live alone? & No & No, Cotton did not live alone. \\
        Can you correct yourself? &  & Cotton lived with her family. \\
        \end{tabular}
        \caption{Conversation with ChatGPT-4}
        \label{tab:week2}
     \end{subtable}
     \caption{Responses to queries based on the paragraph of GPT-3 and GPT-4.}
     \label{tab:temps}
\end{table}

\section{Discussion}
The findings of our study indicate that ChatGPT shows promise in the field of conversational QA, but also reveal the need for enhancements to increase the accuracy and specificity of its responses. To achieve this, future research could consider including external knowledge sources, like knowledge bases, by proposing methods for fact-checking ChatGPT-generated text. It may also be beneficial to investigate alternative approaches for fine-tuning ChatGPT for conversational QA tasks, as this could lead to favorable outcomes.

\section{Conclusion}
In this study, we conducted a thorough comparison between ChatGPT's responses and existing QA corpora to analyze the reliability and suitability of its output for conversational QA tasks. We developed a pipeline that generates large-scale responses and calculated BLEU, ROUGE, and TER scores of ChatGPT's responses. Our results suggest that ChatGPT has great potential for conversational QA tasks, but also highlighted the improvements in the latest GPT-4 model. We hope that our study will contribute to the development of more effective and reliable conversational QA systems based on large-scale language models like ChatGPT.

\bibliographystyle{unsrt}  
\bibliography{main}
\section{Appendix}
\subsection{Evaluation Metrics}
\label{appendix}
\subsubsection{BLEU} The Bilingual Evaluation Understudy (BLEU) Score is a metric used to evaluate the quality of the machine-generated text\cite{10.3115/1073083.1073135}. BLEU measures the similarity between a machine-generated text and a reference text. It is often used in machine translation, where the quality of a translated text is evaluated by comparing it to a reference translation. BLEU calculates a score based on the n-grams (contiguous sequences of words) shared between the machine-generated text and the reference text. To calculate the BLEU score, the system-generated output is compared to one or more reference texts using a modified form of precision. This modified precision takes into account the number of times an n-gram appears in the reference text(s), and the number of times it appears in the system-generated output. The score ranges from 0 to 1, with a higher score indicating a better match between the system-generated output and the reference text(s). 

\begin{equation}
\text{BLEU} = BP \cdot \exp\left(\sum_{n=1}^{N} w_n \log p_n\right)
\end{equation}

where:

\begin{itemize}
\item $BP$ is the brevity penalty,
\item $N$ is the maximum n-gram length,
\item $w_n$ is the weight assigned to the n-gram precision score, and
\item $p_n$ is the n-gram precision score.
\end{itemize}

The brevity penalty term $BP$ is calculated as:

\begin{equation}
\text{BP} = \begin{cases}
\exp(1-\frac{ref}{out}) & \text{if   }  out < ref \\
1 & \text{if   }  out \geq ref
\end{cases}
\end{equation}\\

The $n$-gram precision score $p_n$ is calculated as:

\begin{equation}
p_n = \frac{\sum_{\text{n-grams } g \in \text{output}}\min(count_{ref}(g), count_{output}(g))}{\sum_{\text{n-grams } g \in \text{output}} count_{output}(g)}
\end{equation}\\

where $count_{ref}(g)$ is the number of times n-gram $g$ appears in the reference translation, and $count_{output}(g)$ is the number of times n-gram $g$ appears in the machine-generated translation.

The weights $w_n$ are usually set to $1/N$, where $N$ is the maximum n-gram length.

\subsubsection{ROUGE}Recall-Oriented Understudy for Gisting Evaluation is a set of evaluation metrics used in natural language processing to evaluate the quality of machine-generated summaries\cite{lin-2004-rouge}. ROUGE calculates the overlap between a machine-generated summary and one or more reference summaries. It measures the quality of the summary by comparing the number of overlapping n-grams (contiguous sequences of words) between the machine-generated summary and the reference summary. Like BLEU, ROUGE scores range from 0 to 1, with a higher score indicating a better match between the system-generated summary and the reference summary. ROUGE scores are often reported as F-measure, which takes into account both precision and recall.

\begin{equation}
\text{ROUGE} = \frac{\sum_{r \in R}\sum_{n \in N} count_{match}(n, r)}{\sum_{r \in R}\sum_{n \in N} count_{ref}(n, r)}
\end{equation}\\

where:

\begin{itemize}
\item $R$ is the set of reference summaries,
\item $N$ is the set of n-grams in the machine-generated summary,
\item $count_{match}(n, r)$ is the number of times n-gram $n$ appears in reference summary $r$, and
\item $count_{ref}(n, r)$ is the total number of times n-gram $n$ appears in all reference summaries.
\end{itemize}

\subsubsection{TER}Translation Edit Rate measures the edit distance between the machine-generated translation and the reference translation\cite{agarwal-lavie-2008-meteor}. The edit distance is the number of changes required to transform the machine-generated translation into the reference translation. The changes can include insertions, deletions, substitutions, and shifts. TER calculates the edit distance using a set of pre-defined edit operations that take into account the differences in word order and phrase structure between the machine-generated translation and the reference translation. The TER score is then calculated as the edit distance divided by the length of the reference translation.

\begin{equation}
\text{TER} = \frac{S + D + I}{N}\\  
\end{equation}

where:

\begin{itemize}
\item $S$ is the number of word substitutions required to convert the machine-generated summary into one of the reference summaries,
\item $D$ is the number of words that need to be deleted from the machine-generated summary to match the reference summary, and
\item $I$ is the number of words that need to be inserted into the machine-generated summary to match the reference summary.
\item $N$ is the total number of words in the reference summary.
\end{itemize}

\subsubsection{NIST}The NIST (National Institute of Standards and Technology) score equation is commonly used in evaluating the quality of machine translation outputs\cite{Zhang2004InterpretingBS1}. The equation is as follows:

\begin{equation}
    NIST = \frac{1}{N} \sum_{n=1}^{N} \frac{c_n}{n}
\end{equation}

where:
\begin{itemize}
\item N is the maximum order of $n$-grams used in the evaluation
c\_$n$ is the count of $n$-grams that appear in the machine translation output and reference translation $n$ is the order of the $n$-gram being evaluated ($1$ for unigrams, $2$ for bigrams, etc.)
\end{itemize}

\subsubsection{METEOR}The METEOR (Metric for Evaluation of Translation with Explicit ORdering) score is another popular metric for evaluating the quality of machine translation outputs\cite{banerjee-lavie-2005-meteor}. The equation is as follows:

\begin{equation}
\text{METEOR} = (1-\alpha) \cdot \frac{match_{seg}}{num_{seg}} + \alpha \cdot penalty
\end{equation}

where:
\begin{itemize}
\item $match_{seg}$ is the total unigram, bigram, and trigram matches between the machine translation output and reference translation, weighted by their importance
\item $num_{seg}$ is the number of words in the machine translation output
\item $\alpha$ is a parameter that controls the importance of precision and recall (typically set to 0.5)
\item $penalty$ is a penalty term that accounts for the number of chunks, chunk shifts, and unaligned words between the machine translation output and a reference translation
\end{itemize}

\subsubsection{BARTScore}BARTScore is a metric used to evaluate the quality of the generated text\cite{yuan2021bartscore}. The BARTScore equation is as follows:

\begin{equation}
\text{BARTScore}(x,y) = \exp{{\frac{1}{n} \sum_{i=1}^n \log p_{\text{BART}}(x_i \vert x_{<i}, y) - \frac{1}{m} \sum_{j=1}^m \log p_{\text{BART}}(y_j \vert y_{<j}, x)}}
\end{equation}

where:
\begin{itemize}
\item $x$ is the input text
\item $y$ is the generated output text
\item $p_{\text{BART}}(x_i \vert x_{<i}, y)$ is the conditional probability of predicting the $i^{th}$ word in the input sequence given the previous words in the input sequence and the generated output sequence
\item $p_{\text{BART}}(y_j \vert y_{<j}, x)$ is the conditional probability of predicting the $j^{th}$ word in the generated output sequence given the previous words in the generated output sequence and the input sequence.
\item $n$ is the length of the input sequence.
\item $m$ is the length of the generated output sequence.
\end{itemize}

\end{document}